# Automated urban waterlogging assessment and early warning through a mixture of foundation models

Chenxu Zhang[1], Fuxiang Huang[2], Lei Zhang[1*]

[1*]Chongqing Key Laboratory of Bio-Perception and Multimodal Intelligent Information Processing and the School of Microelectronics and Communication Engineering, Chongqing University, Chongqing, China (*correspondence: leizhang@cqu.edu.cn)

[2]School of Data Science, Lingnan University, Hong Kong, China

## Abstract

With climate change intensifying, urban waterlogging poses an increasingly severe threat to global public safety and infrastructure. However, existing monitoring approaches rely heavily on manual reporting and fail to provide timely and comprehensive assessments. In this study, we present Urban Waterlogging Assessment (UWAssess), a foundation model-driven framework that automatically identifies waterlogged areas in surveillance images and generates structured assessment reports. To address the scarcity of labeled data, we design a semi-supervised fine-tuning strategy and a chain-of-thought (CoT) prompting strategy to unleash the potential of the foundation model for data-scarce downstream tasks. Evaluations on challenging visual benchmarks demonstrate substantial improvements in perception performance. GPT-based evaluations confirm the ability of UWAssess to generate reliable textual reports that accurately describe waterlogging extent, depth, risk and impact. This dual capability enables a shift of waterlogging monitoring from perception to generation, while the collaborative framework of multiple foundation models lays the groundwork for intelligent and scalable systems, supporting urban management, disaster response and climate resilience.

## Introduction

As global climate change continues, extreme weather events are becoming increasingly frequent. Coupled with the challenges posed by rapid urbanization, such as inadequate planning and aging drainage systems, urban waterlogging has been a significant public safety challenge facing numerous cities worldwide. Urban waterlogging refers to the phenomenon where heavy rainfall or prolonged precipitation exceeds a city's drainage capacity, leading to water accumulation disasters within the city. Such events can disrupt urban transportation, cause traffic congestion, damage infrastructure, pose safety risks, and trigger secondary disasters such as sewage overflow and disease transmission. To effectively address these issues and enhance the emergency response capabilities of relevant urban departments, timely and accurate waterlogging monitoring is of critical importance. Traditional monitoring methods often rely on fixed water level sensors[1,2] or manual inspections, which have limitations such as restricted monitoring ranges and high costs. Satellite remote sensing methods[3-5] offer broad coverage, but they lack sufficient spatio-temporal resolution to meet the demands of refined and real-time waterlogging control and disaster reduction in cities, as shown in Fig. 1a. Public surveillance cameras widely distributed across cities can operate continuously around the clock with high reliability, providing a highly promising data source for urban waterlogging monitoring.

In recent years, several vision-based urban waterlogging perception methods[6-14] have been proposed. These methods typically use convolutional neural networks (CNNs), such as FCN[15], Mask RCNN[16], DeepLabv3+[17], PSPNet[18] and SegNet[19], to detect waterlogged regions by extracting visual features from surveillance images. However, the diverse shapes, textures, and turbidity levels of waterlogged regions, combined with the complex lighting conditions and heterogeneous road surfaces in urban environments, often hinder their accuracy and robustness in real-world scenarios. A representative work is LSM-Adapter[20], which introduces a large-small co-adapter paradigm that combines a vision foundation model with a small model such as Mask RCNN[16], SINet[21] and U2Net[28]. This design substantially improves generalization ability and better adapts to diverse urban conditions. Nevertheless, there remains a significant gap between these methods and large-scale deployment in real-world scenes, primarily manifested in the following two aspects. First, these methods are generally designed based on the supervised learning paradigm, requiring a large amount of high-quality annotated data, which is costly and labor-intensive to obtain. Second, existing methods are limited to the visual perception level and cannot provide structured disaster semantic information, offering limited assistance to decision-making and analysis in urban management.

Recently, foundation models represent one of the revolutionary advancements in artificial intelligence (AI), offering new opportunities to address the problems described above. Several typical models, such as CLIP[22], SAM[23], DINO[24], LLaMA[25], DeepSeek-R1[27] and GPT-4[29], are trained on vast amounts of data to learn highly generalizable representations and have demonstrated remarkable capabilities in zero-shot reasoning, transfer learning and multimodal understanding. Despite their success in broad domains, adapting foundation models efficiently and robustly to specialized downstream tasks with data scarcity remains a critical challenge. The fine-tuning process often requires substantial computational resources or large amounts of labeled data, which limits their scalability in data-scarce or resource-constrained environments. Moreover, their potential in domain-specific applications, such as disaster management and urban science, has yet to be fully unlocked, particularly in activating multimodal reasoning that bridges visual perception and structured semantic understanding.

In this work, we introduce an urban waterlogging assessment framework termed UWAssess, which leverages foundation models to achieve both visual perception and textual report generation of urban waterlogging. Our main contributions include the following:

1. A novel multimodal assessment framework for waterlogging warning. We present UWAssess, the first framework that synergistically integrates multiple foundation models for urban waterlogging assessment, as shown in Fig. 1b. This design facilitates a paradigm shift from coarse-grained visual perception to fine-grained report generation, delivering both pixel-level visual perception masks and structured textual reports that capture key attributes of waterlogging states, such as extent, depth, potential risks and impacts.

2. A low-cost semi-supervised fine-tuning strategy for vision foundation models. To reduce reliance on costly annotations, a novel adaptation strategy is designed. The strategy includes (i) a hybrid adaptation module (as presented in Extended Data Fig. 1), integrating adaptation modules with distinct fine-tuning mechanisms through a gating mechanism for flexible feature adaptation; and (ii) a consistency regularization-based semi-supervised learning method, termed S2Match (as presented in Extended Data Fig. 2), employing Structural perturbation and Strong augmentation to fully leverage unlabeled data and promote robust feature learning. This strategy achieves superior generalization across diverse and challenging urban scenes with few labeled data.

3. A training-free prompting strategy for vision-language foundation models. We propose S3CoT, an innovative chain-of-thought prompting strategy that operates without any fine-tuning. By combining Semantic prompts, Spatial prompts and Structural prompts, S3CoT effectively unleashes the inherent capabilities of vision-language foundation models to generate accurate, reliable and well-organized assessment reports, as presented in Extended Data Fig. 3. This approach allows for flexible customization of report content while elegantly avoiding the prohibitive cost without using any specialized report data.

4. Extensive validation and benchmarking. We validate UWAssess across multiple public benchmarks, including a challenging hard subset. Extensive experiments demonstrate that UWAssess achieves excellent performance in both visual perception and textual report generation tasks, confirmed by quantitative metrics and GPT-4 based evaluations. We will release code, models, a curated large-scale unlabeled urban waterlogging dataset (UW-Unlabeled) for training and a textual report dataset (UW-Report) for evaluation to facilitate future research.

This achievement is expected to advance smart city development, laying a crucial foundation for intelligent waterlogging monitoring and emergency response systems in cities, and significantly enhancing a city's capabilities in waterlogging disaster monitoring, early warning, prevention and mitigation.

**Fig. 1: Comparison of different urban waterlogging monitoring systems.**

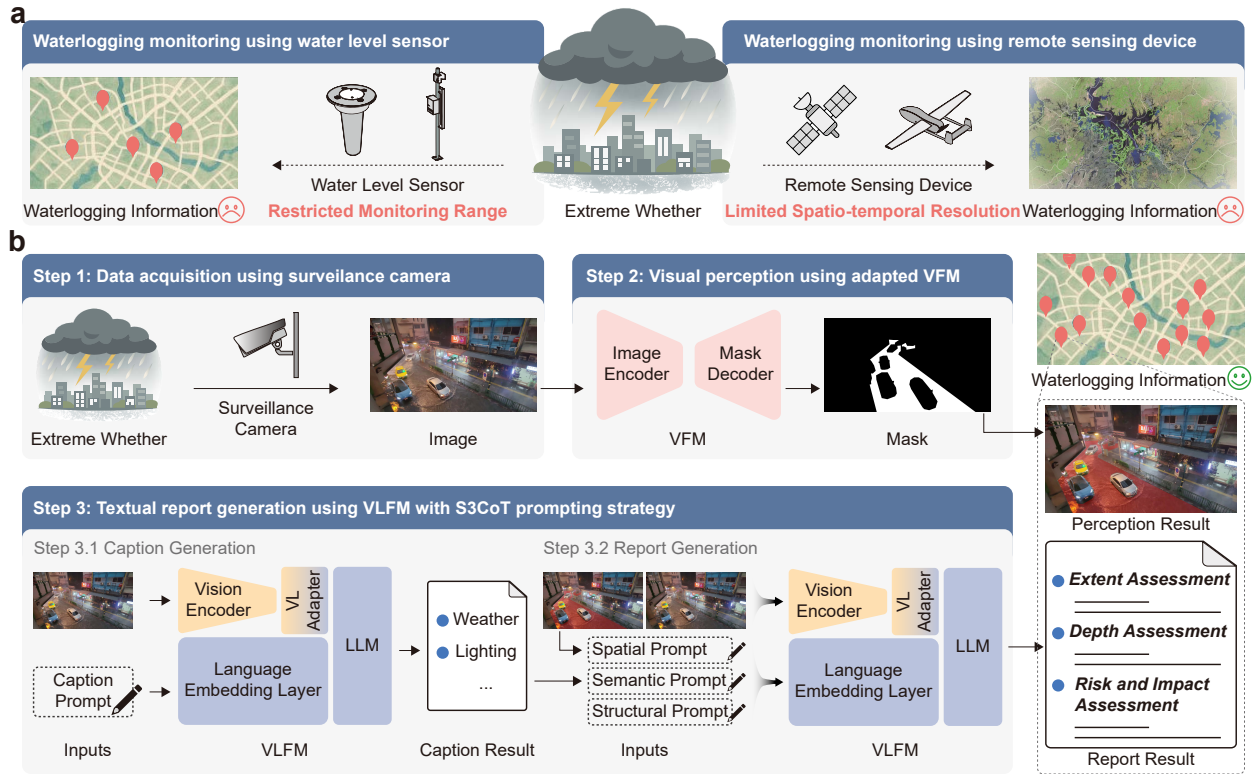

**a** Urban waterlogging monitoring using water level sensors and remote sensing devices. Water level sensor-based approaches have limitations such as high cost and restricted monitoring range. Remote sensing-based approaches have limitations in terms of spatio-temporal resolution. **b** Urban waterlogging events caused by extreme weather are captured by surveillance cameras widely deployed throughout the city and processed by UWAssess. It comprises a vision foundation model (VFM) and a vision-language foundation model (VLFM). The VFM is adapted using a semi-supervised fine-tuning strategy for visual perception of waterlogged areas, while the VLFM is guided by a chain-of-thought prompting strategy, incorporating Semantic, Spatial and Structural prompts, to generate structured textual assessment reports. Together, the visual perceptions and textual reports deliver comprehensive information to city management centers, collaborating with multiple government functional departments in the city, supporting disaster response, emergency dispatch, infrastructure optimization, urban planning and public services, thereby strengthening urban resilience to extreme weather. Detailed designs of the semi-supervised fine-tuning strategy are presented in Extended Data Fig. 1and Extended Data Fig. 2, while detailed designs of the chain-of-thought prompting strategy are presented in Extended Data Fig. 3.

## Results

### Datasets and evaluation protocol

Currently, publicly available urban waterlogging perception datasets remain scarce. The mainstream classic datasets include UWBench[20] and Roadway Flooding[26]. UWBench contains 5,584 annotated images as the training set and 2,124 images as the publicly available V2 test set. The V2 test set is divided into UWBench-All, which includes all test samples, and a challenging subset UWBench-Hard, which contains 1,055 images primarily covering complex cases such as strong light reflection and low illumination. The Roadway Flooding contains 441 images, which is relatively small in scale and thus utilized solely for evaluation in this study.

To support our semi-supervised learning framework, we construct a large-scale unlabeled dataset, named UW-Unlabeled, following a collection strategy similar to that of UWBench. Specifically, we retrieve videos using keywords such as "urban waterlogging", "waterlogged road", and "road flooding". Video frames were sampled at fixed time intervals and then manually filtered to remove redundant or irrelevant content, resulting in a final set of 12,609 high-quality unlabeled images. UW-Unlabeled is combined with the training set of UWBench to form the final training data for our semi-supervised learning framework. We use six metrics to evaluate visual perception performance: precision, recall, Intersection of Union (IoU), Dice, specificity and G-Mean.

For the textual report generation task, we sample 85 representative images from diverse scenes within the UWBench V2 test set as test images. We carefully design textual instructions and invoke the GPT-4[29] Turbo model to generate initial corpus data. These outputs are then manually revised to eliminate hallucinated content and correct factual inaccuracies, ultimately yielding a high-quality textual report dataset for model evaluation, named UW-Report, which includes pairs of test images and reference reports.

To evaluate the quality of the generated textual reports, we adopt a GPT-based scoring method inspired by LLaVA[30]. Based on the corresponding reference report, the GPT-4 Turbo model is instructed to evaluate each report with respect to comprehensiveness and details using carefully designed evaluation prompts. The GPT model not only provides a numerical score from 1 to 10 but also outputs a comprehensive explanation of the evaluation.

## Performance of visual perception

To comprehensively evaluate the visual perception performance of UWAssess, we conduct extensive comparisons with methods employing different network architectures, fine-tuning adaptation strategies, and learning paradigms. The supervised methods are trained using the UWBench training set. The comparison methods include the CNN-based U2Net, the hybrid architecture-based LSM-AdapterU[20] that combines a vision foundation model with U2Net, and two fine-tuning-based methods, i.e., SAM2-Adapter[31] and SAM2-LoRA[32]. Additionally, we design two semi-supervised comparison methods, Semi-SAM2-Adapter and Semi-SAM2-LoRA, which are trained using the latest semi-supervised learning framework, i.e., UniMatch V2[33,34], on our large-scale unlabeled dataset UW-Unlabeled. Fig. 2a presents the radar charts of the six evaluation metrics for each method on the UWBench-All, UWBench-Hard and Roadway Flooding test sets (detailed results are available in Supplementary Table 1), respectively. The results indicate that UWAssess consistently achieves superior overall performance across all test sets. Fig. 2b presents the Precision-Recall (PR) curve comparisons of different methods on the three test sets, demonstrating that UWAssess achieves the optimal break-even point (BEP). Additionally, we sample three test images from each of the three test sets, covering diverse waterlogging appearances, weather conditions, lighting environments, road surface states and camera viewpoints. The corresponding visual perception results are illustrated in Fig. 2c. As is shown, UWAssess delivers stable perception results across various complex urban scenes, demonstrating excellent generalization ability and robustness. Additional visual perception comparisons are shown in Supplementary Fig. 1.

**Fig. 2: Comparisons of visual perception performance.**

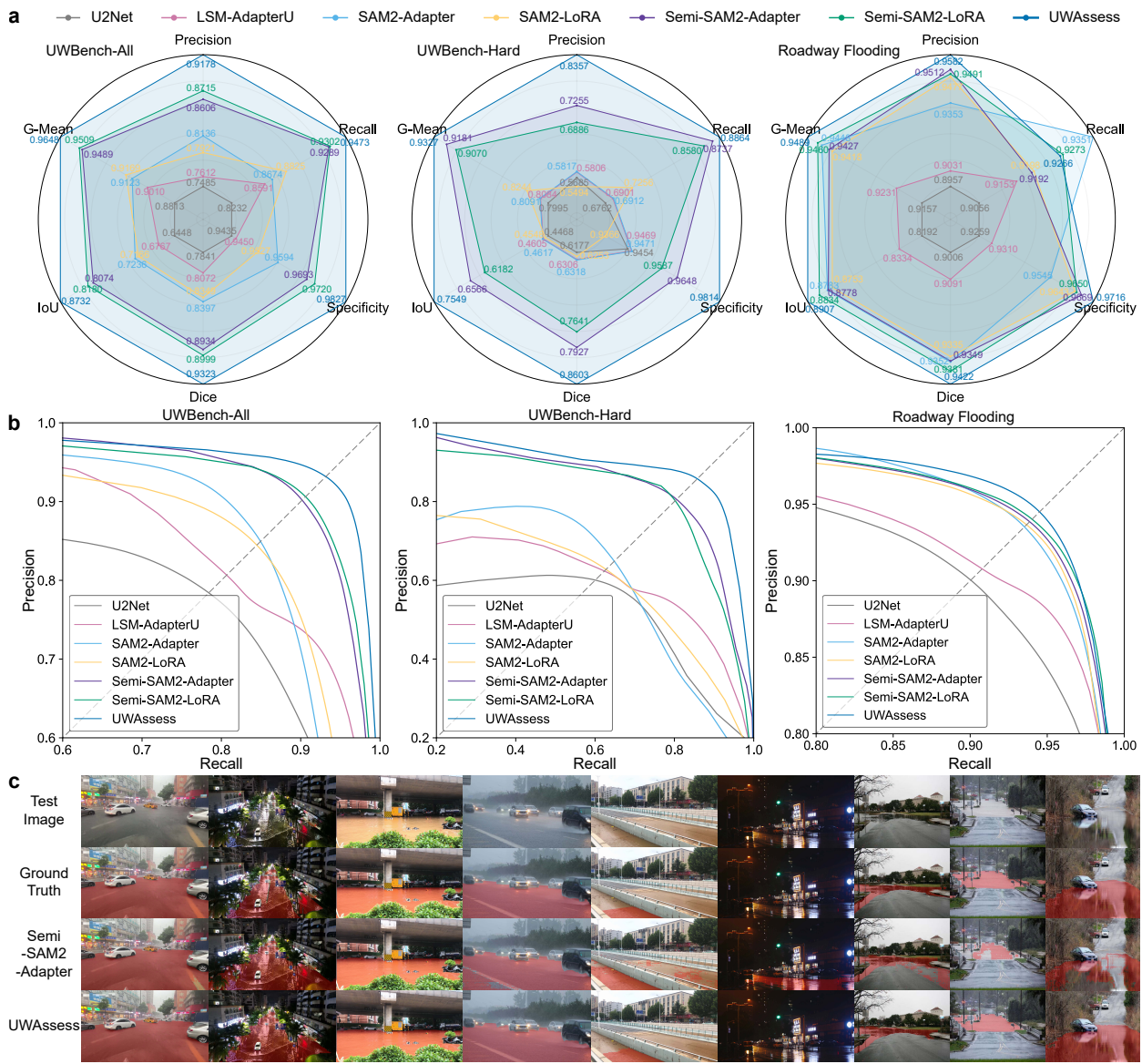

**a** Radar charts comparing six evaluation metrics on three benchmark test sets: UWBench-All, UWBench-Hard and Roadway Flooding. Evaluation metrics include Precision, Recall, IoU, Dice, Specificity and G-Mean. The results demonstrate that UWAssess achieves balanced and competitive performance across all metrics, with clear advantages in comprehensive indicators such as IoU, Dice and G-Mean. Even on UWBench-Hard, that contains challenging cases such as rain, fog, low illumination and reflective surfaces, UWAssess maintains strong stability. While on Roadway Flooding, that differs significantly in scene characteristics, the model still shows competitive generalization. Detailed numerical results are reported in Supplementary Table 1. **b** Precision–Recall (PR) curve comparisons on the three test sets. Diagonal reference lines are included to assist in visualizing break-even points (BEPs). Compared with competing methods, UWAssess exhibits favorable trade-off between precision and recall, achieving the optimal BEPs. **c** Qualitative visual perception results on representative test images. A total of 9 examples are presented, arranged horizontally and evenly divided into three groups, sampled from the three test sets, respectively. Examples cover diverse and complex real-world conditions, including variations in lighting, weather, viewpoint, waterlogging appearance and road surface states. UWAssess accurately and comprehensively identifies the waterlogged regions across all scenarios, avoiding common failure cases such as missing shallow water or misidentifying reflections. These results demonstrate the robustness and strong generalization capability of UWAssess in handling complex urban waterlogging conditions.

## Performance of textual report generation

To evaluate the capability of UWAssess in generating textual assessment reports for urban waterlogging scenarios, we conduct qualitative analyses using UW-Report. Fig. 3 presents comprehensive performance comparisons with and without S3CoT prompting, clearly illustrating the improvements enabled by our design. Fig. 3a illustrates a representative test image and the ground truth. Fig. 3b shows the textual report generated from UWAssess without using the S3CoT prompting strategy and the corresponding scoring report from GPT-4. We observe that, if without structural guidance or contextual information as S3CoT prompting does, the output results have obvious flaws, such as inaccurate or irrelevant statements (highlighted in red), and fail to conduct a critical assessment of the extent or depth of the waterlogging area. The corresponding GPT scoring report (with key statements highlighted in orange) explicitly indicates the lack of detail description and inadequate coverage in the report, and thus outputs a low score of 5. In contrast, Fig. 3c presents a comparison case using the S3CoT prompting strategy. The scoring report, with key statements highlighted in orange, assigns the textual report a score of 8 and highly recognizes its comprehensive, detailed and clear assessment on various risks and impacts. Fig. 3d-f present another set of textual reports for comparison. The comparison results demonstrate that the proposed S3CoT prompting strategy enables UWAssess to produce coherent and informative reports that accurately summarize and assess the extent, depth and associated risks and impacts observed in the input scene.

Fig. 3: Visualization on textual report generation for waterlogging image inputs.

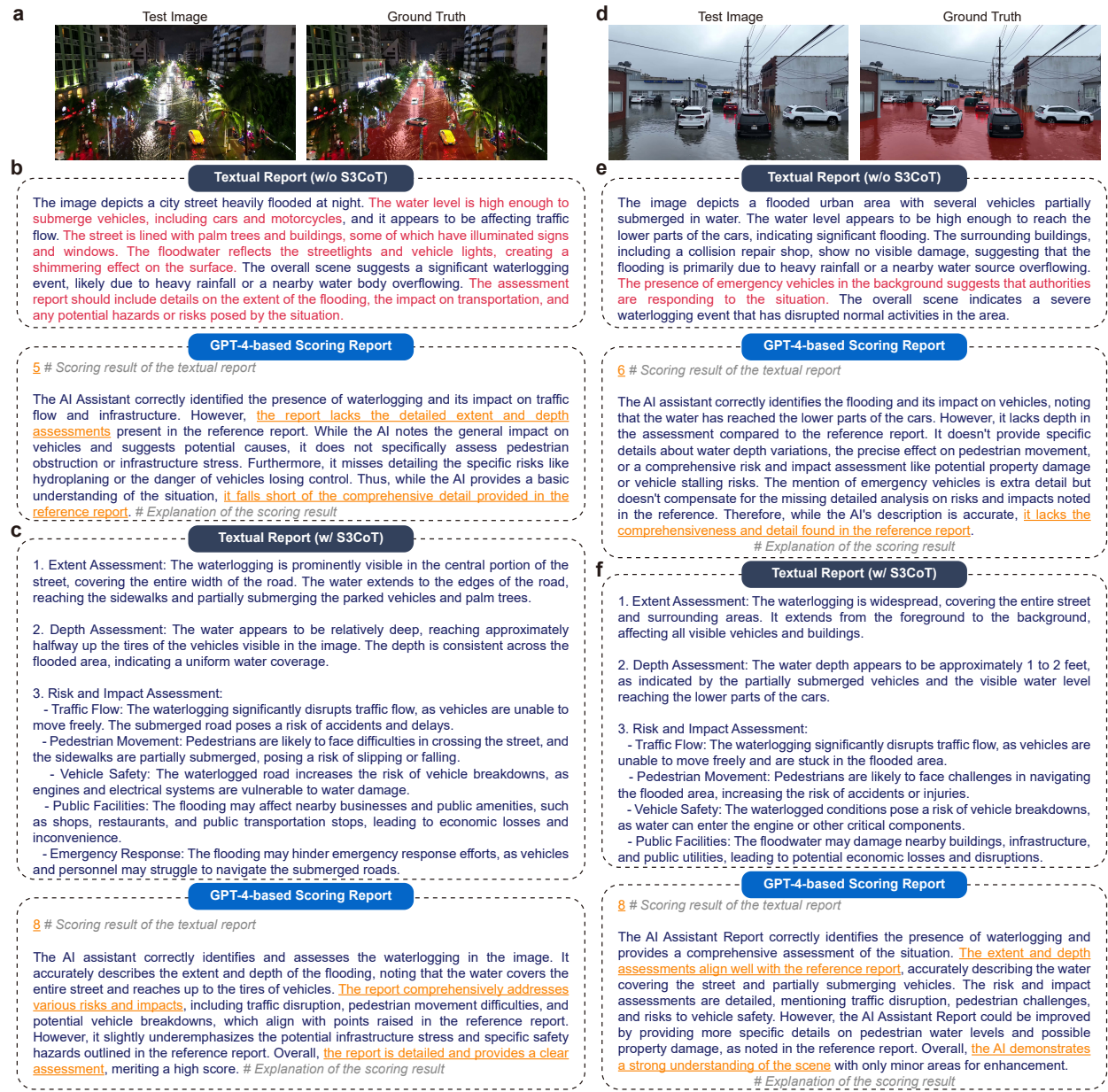

**a, d** Test image and corresponding ground truth mask. **b, e** Textual report generated by UWAssess without using S3CoT prompting strategy, along with the corresponding GPT-4 based scoring report. The scoring report contains an evaluation score to the textual report and provides a comprehensive explanation of the evaluation. The generated textual report lacks structural coherence, omits critical information, and contains irrelevant or incorrect statements (highlighted in red). The scoring report explicitly notes these shortcomings (highlighted in orange), emphasizing insufficient coverage of extent and depth assessment. **c, f** Textual report generated by UWAssess using S3CoT prompting strategy, along with the corresponding GPT-4 based scoring report. The generated textual assessment report is accurate, comprehensive and well-organized, covering important information such as the extent, depth, and potential risks and impacts of waterlogging observed in the input scene. Key positive remarks from the GPT evaluation are highlighted in orange.

## Ablation Studies

**Effectiveness of hybrid adaptation module.** We investigate the impact of fine-tuning different proportions of the encoder layers. Fig. 4a presents the IoU performance on UWBench-All and UWBench-Hard under varying fine-tuning ratios, where each ratio corresponds to tuning a specific number of top encoder layers. The results indicate that increasing the number of fine-tuned encoder layers generally leads to better performance. Even tuning only 25% of the encoder layers yields significant improvements compared to tuning the decoder alone, with IoU gains of 0.0759 on UWBench-All and 0.0696 on UWBench-Hard. These results demonstrate that even partial fine-tuning of the encoder can effectively enhance performance while maintaining parameter efficiency. Detailed performance metrics and trainable parameter size comparisons under different fine-tuning ratios are available in Supplementary Table 2.

To further validate the impact of different fine-tuning modules on feature representation, we visualize attention heatmaps using Grad-CAM[36] on the final neck layer of the encoder. Fig. 4b presents comparative results across various urban scenes. The Frozen setting, where only the decoder is tuned, consistently yields suboptimal results. In contrast, tuning encoder enables the model to better capture the correct waterlogged regions. The stacked approach (Adapter + LoRA) sometimes shows benefits by fusing different adaptation patterns (e.g., the 2$^{nd}$ and 3$^{rd}$ rows in Fig. 4b), but can also result in poor attention (e.g., the 4$^{th}$ row in Fig.

4b). Incorporating a gating unit leads to a more accurate attention, with the model consistently focusing on the waterlogged areas across all scenes, demonstrating stronger adaptability. We further evaluate the temporal robustness of attention patterns by analyzing consecutive frames from the same scene, as shown in Fig. 4c. Despite minimal changes between frames, models using a single adaptation module still exhibit significant fluctuations in attention heatmaps (e.g., the $2^{nd}$ and $3^{rd}$ rows in Fig. 4c), while the model using stacked adaptation modules produces more stable attention patterns (e.g., the $4^{th}$ row in Fig. 4c). In contrast, our proposed hybrid adaptation module helps to consistently focus on target areas, demonstrating both spatial accuracy and temporal robustness across consecutive frames.

We further conduct ablation studies on the proposed hybrid adaptation module, and the experimental results are shown in Fig. 5a-b. Compared to tuning only the decoder, adapting the encoder leads to significantly improved perception performance. However, simply stacking the two adaptation modules (Adapter[31,35] and LoRA[32]) does not yield substantial gains and even degrade the performance. For instance, compared to using only the LoRA adaptation module, the stacked approach yields only marginal improvements in Recall and G-Mean on the UWBench-Hard test set. Using additional gating units enables better integration of different fine-tuning modules. On the UWBench-All test set, the gating unit achieves improvements of 0.0262, 0.0449 and 0.0076 on the three comprehensive metrics of Dice, IoU and G-Mean, respectively. While on the challenging subset UWBench-Hard, the performance gains are even more pronounced, reaching 0.0790, 0.1139 and 0.0099, respectively. Numerical results and trainable parameters are presented in Supplementary Table 3, demonstrating that our hybrid adaptation module achieves notable performance and stability improvements while maintaining parameter efficiency.

**Effectiveness of the semi-supervised learning method, i.e., S2Match.** The impact on model performance of the confidence threshold in strong-to-strong consistency (SC) for selecting reliable pseudo-labeled data is visualized. The experimental results on the UWBench-All and UWBench-Hard are shown in Fig. 4d-e, respectively (detailed results are available in Supplementary Table 4). We test four different confidence threshold settings: 0.5, 0.65, 0.8 and 0.95. When the threshold is set to 0.95, only pixels with very high confidence participate the consistency loss calculation, while when the threshold is set to 0.5, all pixels are utilized for the consistency loss calculation. The best overall performance is observed at a threshold of 0.8, indicating that moderately relaxing the threshold to select more pseudo-labeled data can bring performance gains. However, when the threshold is further decreased to 0.65 and 0.5, the overall performance gradually declines, indicating that an over-relaxed confidence threshold introduces too many noisy pseudo labels, which can interfere with model training and ultimately affect model performance.

To investigate the impact of the quantity of unlabeled training data on semi-supervised learning performance, we randomly sample 100 test images from three test sets and show box plots of IoU performance, as shown in Fig. 4f. Based on the scale of labeled training data (5,584 images), we define four settings for the additional unlabeled training data: 0% (the same size as the labeled data, i.e., 5,584 images), 25% (7,340 images), 50% (9,097 images), and 100% (using all 12,609 images). As shown in the figure, as the proportion of additional unlabeled training data increases, both the median and mean IoU (marked with diamond points) exhibit an upward trend, indicating that the model's perception capability improves as the number of unlabeled training data increases. Notably, under the 100% unlabeled data setting, the IoU distribution becomes more compact, with narrower interquartile ranges and fewer outliers. This suggests improved prediction consistency and reduced performance fluctuation across samples. In summary, incorporating more unlabeled training data not only enhances the model's perception performance but also effectively improves its prediction stability across diverse urban waterlogging scenarios.

Fig. 5c-d show the ablation experiment results on the strong-to-strong consistency (SC) and scale-wise stochastic depth (SD) components. Detailed numerical results are reported in Supplementary Table 5. On the UWBench-All test set, using the SC component alone can bring gains of 0.0216, 0.0354 and 0.0062 in the Dice, IoU and G-Mean metrics, respectively, while on the UWBench-Hard test set, it can bring gains of 0.0202, 0.0270 and 0.0031, respectively. However, using the SD component alone results in performance degradation on multiple key metrics on the UWBench-Hard. When SC and SD are used jointly, the model benefits from their complementary effects and shows consistent performance improvement across both test sets. Specifically, on the UWBench-All test set, the Dice, IoU and G-Mean performance metrics are improved by 0.0457, 0.0768 and 0.0173, respectively, while on the UWBench-Hard test set, they are improved by 0.0946, 0.1346 and 0.0345, respectively. These results indicate that while exploring the extended feature perturbation space is promising, it must be paired with a reasonable consistency constraint, or it may negatively disturb model training and lead to performance degradation.

**Effectiveness of S3CoT prompting strategy.** We further conduct ablation studies to validate the effectiveness of the proposed S3CoT prompting strategy. Fig. 5e shows violin plots of GPT scoring results for 85 test samples under various ablation settings. Without any prompt design, the vision-language foundation model directly generates textual reports using input images and instructions only, yielding the lowest average score of 5.13, with disorganized text structure and insufficient information coverage. Using the structural prompt significantly improve clarity and text organization, increasing the average score to 6.93. The effectiveness of the structural prompt is demonstrated in Supplementary Fig. 2. Adding the semantic prompt and the spatial prompt further improves the scores by 0.11 and 0.22, respectively. The complete S3CoT prompting strategy, combining all three prompt components, achieve the best performance with an average score of 7.27. The generated reports perform the best in terms of structure accuracy and information completeness. The above results indicate that a well-designed chain-of-thought prompting strategy can effectively unleash the understanding and expression capabilities of a vision-language foundation model, significantly improving its text generation quality in complex urban scenarios.

**Fig. 4: Analysis of hybrid adaptation module and S2Match.**

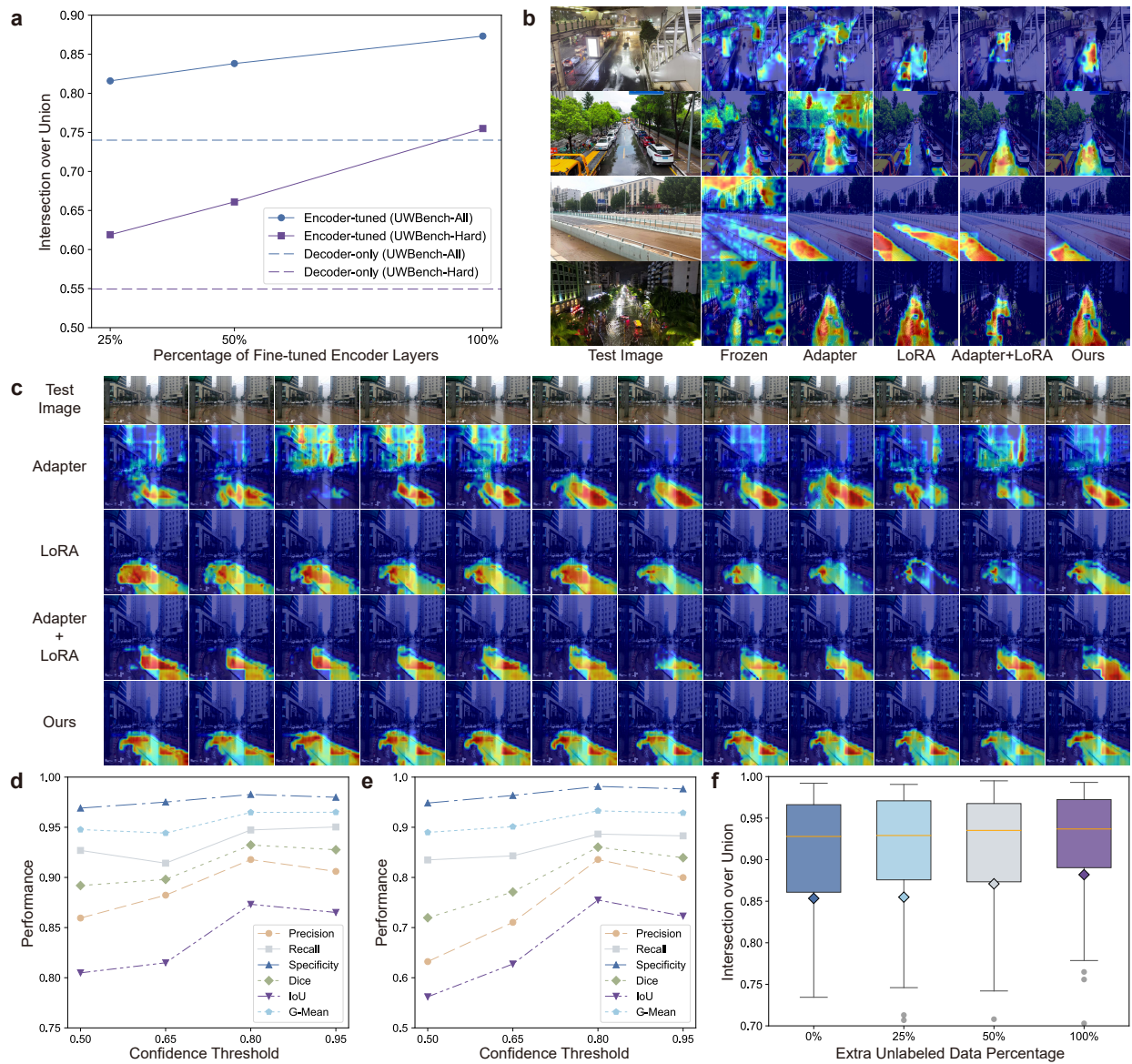

**a** IoU performance on UWBench-All and UWBench-Hard under different encoder fine-tuning ratios. Each ratio corresponds to tuning a specific number of top encoder layers. Increasing the proportion of fine-tuned encoder layers generally improves performance, and even tuning only 25% of the encoder provides notable gains over decoder-only tuning, demonstrating the efficiency of partial adaptation. Detailed results are provided in Supplementary Table 2. **b** Grad-CAM heatmaps visualized on the final neck layer of the encoder under different adaptation settings. Compared with the frozen encoder and single adaptation module baselines (Adapter or LoRA only), our hybrid adaptation module consistently achieves more accurate attention on waterlogged regions, demonstrating its superior adaptation capacity across varied urban scenes. **c** Temporal robustness analysis on consecutive frames from the same scene. Despite minimal differences between frames, models using single adaptation modules show unstable and shifting attention, while our hybrid adaptation module maintains stable and accurate attention patterns. **d-e** Effects of confidence threshold settings in the strong-to-strong consistency (SC) component on UWBench-All (d) and UWBench-Hard (e). A threshold of 0.8 achieves the best balance, leveraging more unlabeled pixels while avoiding excessive noise. Further relaxation to 0.65 or 0.5 degrades performance due to noisy pseudo labels. Detailed results are reported in Supplementary Table 4. **f** Box plots of IoU distributions on 100 randomly sampled test images from the three datasets under four different proportions of additional unlabeled training data. Medians, quartiles and outliers are shown, with diamond markers indicating mean values. The results demonstrate that increasing the amount of unlabeled data improves both average performance and stability.

**Fig. 5: Ablation studies on semi-supervised fine-tuning strategy and S3CoT prompting strategy.**

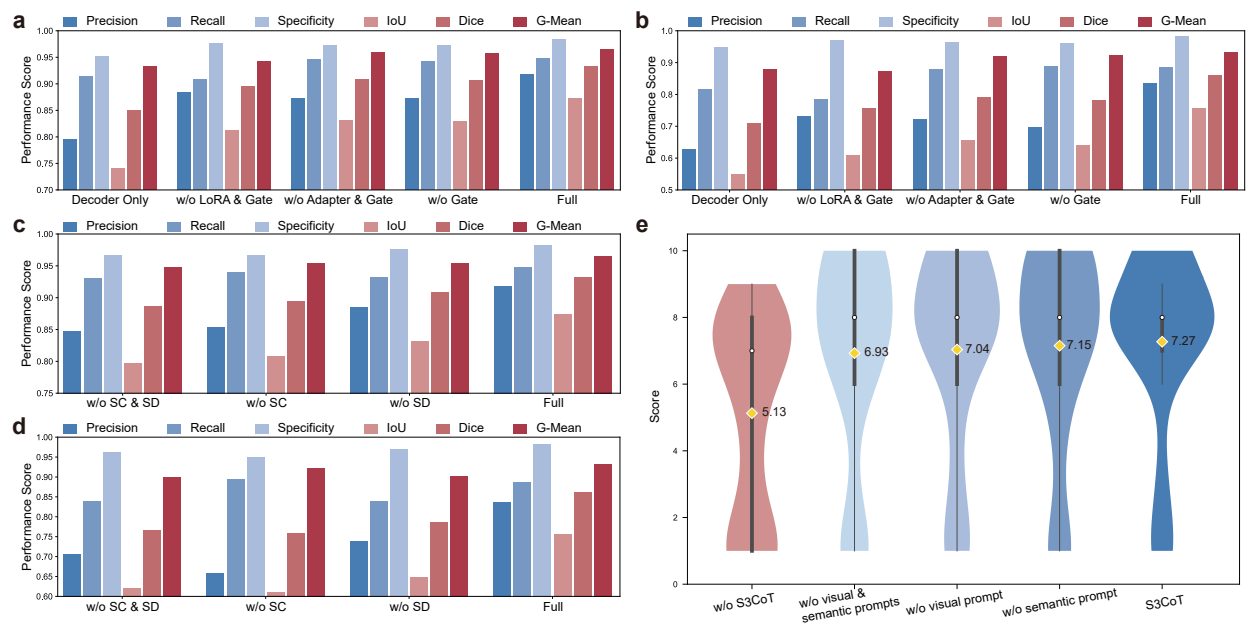

**a-b** Ablation studies of the hybrid adaptation module on UWBench-All and UWBench-Hard, respectively. Adapting the encoder achieves better performance than tuning the decoder alone, while additional gating units enable better integration of different adaptation modules to boost performance. Detailed numerical results are presented in Supplementary Table 3. **c-d** Ablation studies of the S2Match semi-supervised learning method on UWBench-All and UWBench-Hard, respectively. The combination of strong-to-strong consistency (SC) and scale-wise stochastic depth (SD) leads to optimal performance. Detailed numerical results are presented in Supplementary Table 5. **e** Ablation studies of the S3CoT prompting strategy on UW-Report. Diamond markers indicate average scores. The optimal overall performance is achieved using the complete S3CoT prompting strategy.

## Discussion

In this study, we present UWAssess, a comprehensive urban waterlogging assessment framework that leverages the widely deployed surveillance cameras in cities to achieve automatic and intelligent waterlogging monitoring. By integrating the vision foundation model SAM2[37] and the vision-language foundation model DeepSeek-VL2[38], UWAssess supports both visual perception and textual report generation, providing auxiliary information for urban management departments and enhancing emergency dispatch, disaster prevention and mitigation efficiency.

Extensive experiments demonstrate that UWAssess exhibits strong generalization and robustness, delivering excellent visual perception and textual report generation capabilities across diverse and complex urban environments. The gains arise from three methodological innovations. First, the hybrid adaptation module effectively balances the complementary strengths of Adapter and LoRA. Adapter injects task-specific features into intermediate representations between encoder layers, LoRA fine-tunes the self-attention modules inside encoder layers by learning low-rank weight matrices. The gating unit harmonizes the two adaptation modules, yielding more stable and accurate features. Second, the S2Match substantially enhances generalization capability and robustness with a lower cost. Combining the consistency of strong augmentation streams with scale-wise structural perturbation, S2Match enables the vision foundation model to extract perturbation-invariant features even with very few labeled data. Finally, the S3CoT prompting strategy unleashes the inherent capabilities of the vision-language foundation model without fine-tuning.

Combining multiple prompts with different sources (e.g., VFM, pre-caption from a VLM, manual design, etc.) and functions (spatial position, contextual information, format guidance, etc.), S3CoT enables the generation of accurate, detailed and comprehensive textual reports.

Beyond methodological contributions, UWAssess fully unleashes the power of foundation models and achieve demonstrative application in urban science and disaster management. Surveillance cameras originally deployed for traffic and security purposes can be repurposed as large-scale and real-time environmental sensors when paired with foundation models. This enables cost-effective disaster monitoring by transforming existing infrastructure into intelligent sensing networks. Moreover, by integrating multiple foundation models, the framework transcends pixel-level perception and moves toward structured semantic assessment, providing richer and valuable insights for decision-makers, which aligns with the emerging vision of urban digital twins and AI-driven city brains. In addition, UWAssess demonstrates the collaborative potential of foundation models in urban applications and lays the groundwork for scalable mixture of foundation model (MOF) systems, which could serve as core components for digital urban management and smart emergency response. In the context of increasingly severe climate challenges facing global cities, such comprehensive frameworks play a critical role in enhancing urban disaster resilience and early warning capabilities. In summary, the emergence of UWAssess contributes to improving people's well-being, promotes the development of intelligent and collaborative precision governance models, and provides a path toward sustainable development and public safety enhancement.

Although UWAssess offers numerous advantages, several aspects still need further exploration. The integration of a large vision-language foundation model inevitably increases computational demand. While its modular design supports flexible deployment tailored to specific tasks, scaling to a citywide and real-time system will require continued innovation in model compression and distributed inference. Moreover, as global climate dynamics evolve, rainfall patterns may shift beyond existing data distributions. Due to the essence of ever-changing world, its effectiveness in long-term or life-long deployment is an open challenge for the field at large. Future research should focus on integrating UWAssess with meteorological data streams, hydrological simulations, and continuous learning paradigms to form a closed-loop system integrating perception, generation and decision-making.

In essence, UWAssess not only advances the frontier of multimodal foundation model adaptation but also redefines how AI can serve humanity in managing climate resilience and safeguarding cities. By uniting perception and cognition within a single intelligent framework, it paves the way for a new generation of AI-driven urban governance systems, capable of transforming passive observation into proactive and collective intelligence for the benefit of society.

## Methods

### Overview of UWAssess

As illustrated in Fig. 1, UWAssess consists of two core models: a vision foundation model and a vision-language foundation model, which are responsible for visual perception and textual report generation, respectively, jointly enabling a real-time and comprehensive assessment of urban waterlogging.

**Visual perception.** The vision foundation model adopted is Segment Anything Model 2 (SAM2). To address the scarcity of labeled image data for the urban waterlogging perception task in adverse conditions, we propose a semi-supervised fine-tuning strategy to achieve efficient and robust adaptation, incorporating two key designs: the hybrid adaptation module and the S2Match for semi-supervised learning.

Considering that different adaptation modules possess distinct fine-tuning mechanisms and adaptation capabilities, we propose a hybrid adaptation module to achieve complementary advantages between different adaptation modules. The proposed hybrid adaptation module includes two different adaptation modules, LoRA and Adapter, and a gating unit is designed to achieve a flexible and effective integration of them.

UniMatch[33] is a classic semi-supervised learning method based on consistency regularization, which expands the perturbation space by introducing dual strong augmentation views. UniMatch V2[34] decouples the learning process of dual strong augmentation streams through a channel-wise complementary feature augmentation approach, delivering superior semi-supervised learning performance on Transformer-based architectures. However, these methods employ only Dropout-based[40] feature augmentation, resulting in limited perturbation effects. Secondly, they neglect to leverage predictions from the strong augmentation streams, resulting in inadequate utilization of unlabeled data. Building upon UniMatch V2, we propose a semi-supervised method, termed S2Match, which further explores the structural perturbations and consistency between the two strong augmentation streams to fully exploit unlabeled data. First, the structural perturbation is explored through a scale-wise stochastic depth[41] (SD) mechanism, where encoder layers are randomly skipped during training, thereby expanding the diversity of feature-level perturbations. Second, strong-to-strong consistency (SC) is enforced between two strong augmentation streams, extending the original weak-to-strong paradigm[33,34,39] and guiding the model to capture robust perturbation-invariant representations.

**Textual reporting.** The visual-language foundation model used is DeepSeek-VL2. To address the challenges of preparing large-scale textual datasets and to achieve structured and accurate textual reports, we design a training-free semantic-spatial-structural chain-of-thought prompting strategy, termed S3CoT, to guide the model in generating textual reports. The semantic prompt, derived from an auxiliary captioning task, provides coarse-grained semantic information such as weather and lighting to support scene understanding. The spatial prompt, based on the visual perception result, offers fine-grained spatial information to locate waterlogged areas accurately. The structural prompt serves to guide the format and clarify the key points of analysis. These prompt components work together to effectively unleash the potential of the vision-language foundation model, enhancing the accuracy, completeness and readability of textual reports.

## Vision foundation model and semi-supervised fine-tuning strategy

**Vision foundation model.** SAM2 is an upgraded version of the Segment Anything Model (SAM)[23] developed by Meta AI, serving as a promptable segmentation framework. SAM2 comprises an image encoder, memory attention, prompt encoder, mask decoder, memory encoder and memory bank, enabling point, box and mask prompts for both image and video segmentation. To enable automatic urban waterlogging detection, we fine-tune only the image encoder and mask decoder, while the prompt encoder and memory-related modules are disabled for this task. The image encoder in SAM2 is based on the Hiera[42,43] encoder, and according to the model size of the encoder, SAM2 includes variants T, S, B+ and L.

To facilitate subsequent method descriptions, we outline its feature extraction pipeline here. The encoder comprises four feature extraction stages, progressively downsampling the features to produce multi-scale representations. The third stage serves as the primary feature extraction stage, containing the largest number of layers dedicated to rich semantic modeling, while other stages mainly provide multi-scale contexts to enhance spatial perception. Finally, a feature pyramid network (FPN)[44] fuses high-level features from the third and fourth stages for improved feature representation. Given an input image $x$, the feature extraction process can be represented as follows:

$$f_1 = E_1(x), f_2 = E_2(f_1), f_3 = E_3(f_2), f_4 = E_4(f_3), \tag{1}$$

$$[\hat{f}_1, \hat{f}_2, \hat{f}_3, \hat{f}_4] = \text{Neck}(f_1, f_2, f_3, f_4), \tag{2}$$

$$\hat{f}_3^{fuse} = \hat{f}_3 + \text{Upsampling}(\hat{f}_4), \tag{3}$$

where $\{E_i\}_{i=1}^4$ denote the feature extraction stages in the encoder, $\text{Neck}(\cdot)$ denotes the bottleneck block that adjusts the number of channels for the multi-scale features using convolution layers, and $\text{Upsampling}(\cdot)$ denotes upsampling interpolation. For convenience, the complete feature extraction is defined by using the image encoder $E(\cdot)$.

**Hybrid adaptation module.** The Parameter Efficient Fine-Tuning (PEFT) method adjusts only a small number of parameters in the model, significantly reducing memory usage and computational overhead. Additionally, freezing most of the pre-trained parameters helps mitigate overfitting issues, making it particularly suitable for real-world scenarios with scarce labeled data.

In this work, we adopt two representative PEFT methods: Adapter and LoRA, each adjusting different components of the encoder. The Adapter extracts task-specific information from the input image (e.g., high-frequency details beneficial for segmentation) and injects such information into each encoder layer via a combination of unshared and layer-shared multi-layer perceptrons (MLPs). In contrast, the LoRA modifies the multi-head self-attention layers by adding trainable low-rank matrices alongside the original attention weights, enabling fine-grained control over the attention distribution. Given that the two methods adapt distinct parts of the encoder and exhibit complementary advantages, we propose a hybrid adaptation module that integrates both mechanisms, as shown in the Extended Data Fig. 1. To coordinate the two adaptation methods more effectively, we introduce a gating unit that adaptively learns to balance the contributions of Adapter and LoRA for each layer, enabling flexible and effective fusion of their fine-tuning effects.

**S2Match semi-supervised method.** Our semi-supervised learning method is developed based on the UniMatch V2. It belongs to the consistency regularization based semi-supervised learning method. Extended Data Fig. 2 illustrates the proposed method, and the following is a detailed description.

The training dataset consists of a labeled dataset $\mathcal{D}^l = \{(x_i^l, y_i^l)\}$, and an unlabeled dataset $\mathcal{D}^u = \{x_i^u\}$, where $x_i$ represents the $i$-th image and $y_i$ is the corresponding ground truth mask. Each mini-batch contains $B_l$ labeled images and $B_u$ unlabeled images. For labeled data, the model is supervised by the provided labels. The loss function for labeled data can be expressed as follows:

$$\mathcal{L}^l = \frac{1}{B_l} \sum_{i=1}^{B_l} \mathrm{H}(p_i^l, y_i^l), \tag{4}$$

where $p_i^l$ is the prediction probability of the $i$-th labeled image, $y_i^l$ is the corresponding ground truth mask, and $\mathrm{H}(\cdot)$ denotes the binary cross-entropy loss.

We use both image-level and feature-level perturbations for unlabeled data to explore the vast perturbation space. For an unlabeled image $x^u$, we obtain the corresponding weakly augmented images $x^w$ through weak augmentation method $\mathcal{A}^w$, and further obtain the corresponding two strongly augmented images $x^{s_1}$ and $x^{s_2}$ through strong augmentation method $\mathcal{A}^s$. The augmentation and corresponding prediction process can be represented as:

$$x^w = \mathcal{A}^w(x^u), \ x^{s_1} = \mathcal{A}^s(x^w), \ x^{s_2} = \mathcal{A}^s(x^w), \tag{5}$$

$$p^w = F_{\theta^t}(x^w), \ p^{s_1} = F_{\theta^s}(x^{s_1}), \ p^{s_2} = F_{\theta^s}(x^{s_2}), \tag{6}$$

where $F_{\theta^s}$ is the student model with model parameters $\theta^s$, and $F_{\theta^t}$ is the teacher model obtained using exponential moving average (EMA)[45], with its model parameters $\theta^t$ updated as follows:

$$\theta^t \leftarrow \gamma \times \theta^t + (1-\gamma) \times \theta^s, \tag{7}$$

where $\gamma$ is dynamically updated as $\min(1 - \frac{1}{iter+1}, 0.996)$. Note that $\min(\cdot)$ denotes the minimum operator and $iter$ represents the index of the current iteration. The prediction $p^w$ of the weak augmentation image $x^w$ serves as a pseudo label and is binarized with a threshold of 0.5 to obtain the binary mask $\hat{p}^w$.

For strong augmented images, we impose two feature perturbations: scale-wise stochastic depth (SD) and channel-wise complementary Dropout (CD). Given a strong augmented image $x^s$, the prediction process can be represented as:

$$p^s = D(\mathcal{F}(E_{SD}(x^s))), \tag{8}$$

where $E_{SD}(\cdot)$ denotes the feature extraction function based on the SD module, $\mathcal{F}(\cdot)$ denotes the CD module, and $D(\cdot)$ denotes the mask decoder. The SD module enables randomly skipping the lowest-resolution path in the encoder, i.e., the 4$^{th}$ feature extraction stage, creating structural feature perturbations. In $E_{SD}(\cdot)$, the feature fusion process between the 3$^{rd}$ and 4$^{th}$ stages is modified as follows:

$$\hat{f}_3^{fuse} = \hat{f}_3 + \frac{b}{1-p}\text{Upsampling}(\hat{f}_4), \tag{9}$$

where $p$ is the probability of random skipping, $b \sim \text{Bernoulli}(p)$. It should be noted that this module is also enabled for labeled images. The CD module $\mathcal{F}$ processes the extracted multi-scale features at the channel level, respectively. Given features $\hat{f}^{s_1}$ and $\hat{f}^{s_2}$ obtained from $E_{SD}(x^{s_1})$ and $E_{SD}(x^{s_2})$, we sample a binary dropout mask $M$ of the same dimension from a binomial distribution with a probability of 0.5. Half of the channels in $M$ are set to 1, while the other half are set to 0. Using $M$, we perform channel-wise complementary Dropout operations on the features:

$$\hat{f}^{s_1} \leftarrow \hat{f}^{s_1} \odot M \times 2, \tag{10}$$

$$\hat{f}^{s_2} \leftarrow \hat{f}^{s_2} \odot (1-M) \times 2. \tag{11}$$

Predictions of the two strong augmented images are then obtained as:

$$p^{s_1} = D(\hat{f}_1^{s_1}, \hat{f}_2^{s_1}, \hat{f}_3^{s_1\_fuse}),\ p^{s_2} = D(\hat{f}_1^{s_2}, \hat{f}_2^{s_2}, \hat{f}_3^{s_2\_fuse}), \tag{12}$$

In addition to the standard weak-to-strong consistency, we impose strong-to-strong consistency to promote perturbation-invariant feature learning and make full use of unlabeled data. The loss function $\mathcal{L}^u$ for unlabeled data comprises weak-to-strong consistency loss $\mathcal{L}_{ws}^u$ and strong-to-strong consistency loss $\mathcal{L}_{ss}^u$, formulated as follows:

$$\mathcal{L}_{ws}^u = \frac{1}{4B_u}\sum_{i=1}^{B_u} \mathbb{I}(p_i^w \geq \tau \vee p_i^w \leq 1-\tau) \cdot (\mathrm{H}(p_i^{s_1}, \hat{p}_i^w) + \mathrm{H}(p_i^{s_2}, \hat{p}_i^w)), \tag{13}$$

$$\mathcal{L}_{ss}^u = \frac{1}{4B_u}\sum_{i=1}^{B_u} \mathbb{I}(p_i^w \geq \tau_s \vee p_i^w \leq 1-\tau_s) \cdot (\mathrm{H}(p_i^{s_1}, \hat{p}_i^{s_2}) + \mathrm{H}(p_i^{s_2}, \hat{p}_i^{s_1})), \tag{14}$$

$$\mathcal{L}^u = \mathcal{L}_{ws}^u + \mathcal{L}_{ss}^u, \tag{15}$$

where ∨ denotes the logical OR operator and $\mathbb{I}(\cdot)$ is an indicator function that outputs 0 or 1 based on its internal condition. It mitigates the negative effect of noisy pseudo labels using a confidence threshold $\tau = 0.95$ for the weak augmented images (following UniMatch V2) and a slightly relaxed confidence threshold $\tau_s = 0.8$ for strong augmented images to include more reliable pixels. The binarized pseudo labels $\hat{p}^{s1}$, $\hat{p}^{s2}$ are obtained with a threshold of 0.5, analogous to $\hat{p}^w$, and used for cross-supervision between the two strong-augmented branches. The total loss for a mini-batch is the combination of the supervised objective in equation (4) and the unsupervised objective in equation (15), computed by:

$$\mathcal{L} = \mathcal{L}^l + \lambda \mathcal{L}^u, \tag{16}$$

where $\lambda$ is used to balance the effects of unlabeled data, with a default setting of 1.

## Vision-language foundation model and chain-of-thought prompting strategy

**Vision-language foundation model.** DeepSeek-VL2 is an upgraded version of the previous DeepSeek-VL[46]. It comprises a vision encoder, a vision-language adapter, a language embedding layer and a large language model (LLM), which supports a broad spectrum of vision-language tasks such as visual question answering (VQA) and optical character recognition (OCR). The vision encoder is SigLIP-SO400M-384[47], combined with a dynamic tiling strategy to handle high-resolution images with diverse aspect ratios, mitigating distortions from naive resizing. The vision-language adapter consists of MLP of two layers, projecting visual tokens into the embedding space of the language model. The language embedding layer tokenizes and encodes textual inputs. The LLM model is DeepSeekMoE[48,49], which leverages Multi-head Latent Attention[50] to improve throughput and employs a MoE architecture to achieve efficient inference. Three variants are available: DeepSeek-VL2-Tiny, DeepSeek-VL2-Small and DeepSeek-VL2. Given an input image $I$ and a textual prompt $P$, a text response $G$ can be generated as follows:

$$G = LLM_\varphi(V_\omega(I), L(P)), \tag{17}$$

where $V_\omega(\cdot)$ denotes the combination of the vision encoder and the vision-language adapter, with model parameters $\omega$, $L(\cdot)$ denotes the language embedding layer, and $LLM_\varphi(\cdot)$ denotes the large language model with model parameters $\varphi$.

**S3CoT prompting strategy.** The proposed S3CoT prompting strategy contains three complementary prompt information to guide vision-language foundation model for generating accurate, comprehensive and well-organized textual reports. Specifically, the semantic prompt provides coarse-grained scene semantics such as weather and lighting, the spatial prompt provides fine-grained localization derived from visual perception, and the structural prompt presents the format and key aspects of the generated textual reports content. After completing the visual perception task, the S3CoT prompting strategy is used to generate a textual report in two steps.

Step 1: Capturing scene semantic information via image caption. Given an input image $I$, we use a designed captioning task instruction $T_1$ to obtain a scene-level caption $C$ as follows:

$$P_I = O_I(I), \ P_{cap} = O_{cap}(T_1)), \tag{18}$$

$$C = LLM_\varphi(V_\omega(I), L(P_I, P_{cap})), \tag{19}$$

where $P_I$ and $P_{cap}$ are used to clearly indicate the input image and the caption task, respectively, $O_I(\cdot)$ and $O_{cap}(\cdot)$ represent the operations of adjusting the format and content of the corresponding inputs (i.e., image and text). For example, the $P_I$ processed by $O_I(\cdot)$ is "Image: <image>", where <image> is a special token used to represent an image in the DeepSeek-VL2 model.

Step 2: Generating a structured textual report by fusing semantic, spatial and structural prompts. Given the caption $C$, the visual perception result $S$ and the report generation task instruction $T_2$, the textual report $R$ is generated as follows:

$$P_{sem} = O_{sem}(C),\ P_{spa} = O_{spa}(S),\ P_{str} = O_{str}(T_2), \tag{20}$$

$$R = LLM_\varphi(V_\omega(I, S), L(P_I, P_{sem}, P_{spa}, P_{str})), \tag{21}$$

where $P_{sem}$, $P_{spa}$ and $P_{str}$ are semantic prompt, spatial prompt, and structural prompt, respectively, while $O_{sem}(\cdot)$, $O_{spa}(\cdot)$ and $O_{str}(\cdot)$ are corresponding operators. Detailed task instructions design and operations are shown in Extended Data Fig. 3.

## Textual report test dataset construction and performance evaluation

We use the GPT-4 Turbo API to generate initial text report corpora, which are then manually revised to remove hallucinated content and correct factual errors, and then develop the UW-Report dataset. During corpus generation, we design text instructions as the System Message and upload images as the User Message. For performance evaluation, we also use GPT-4 Turbo. Given the test image, reference report, generated report and evaluation requirements, the GPT-4 Turbo provide scores and corresponding explanations. The test dataset construction process and the performance evaluation process are shown in Extended Data Fig. 4 and Extended Data Fig. 5, respectively.

## Implementation details

For both the vision foundation model and the vision-language foundation model, we use the Small version. In semi-supervised fine-tuning, each mini-batch contains 2 labeled images and 2 unlabeled images. The weak augmentation $\mathcal{A}^w$ includes random resizing (scaling from 0.75 to 1.25), random cropping and random horizontal flipping with probability 0.5. The strong augmentation $\mathcal{A}^s$ includes color jittering with probability 0.8, grayscaling with probability 0.1 and random Gaussian blurring with probability 0.5. In the color jittering operation, the random jittering range for brightness, contrast and saturation is [0.5, 1.5], and for hue is [-0.25, 0.25]. We use the AdamW[51] optimizer with an initial learning rate of 2e-4, adjusted using a poly learning rate scheduler. The total training epochs are set as 30. The framework is implemented using PyTorch. Training and evaluation of the visual perception model are conducted on one V100 (32GB) GPU, and the textual report evaluation is conducted on one A800 (80GB) GPU.

## Data Availability

All data will be publicly available upon acceptance.

## Code Availability

The code will be available upon acceptance.

## Acknowledgements

This work was partially supported by National Natural Science Fund of China under Grants 92570110 and 62271090, Chongqing Natural Science Fund under Grant CSTB2024NSCQ-JQX0038, National Key R&D Program of China under Grant 2021YFB3100800 and National Youth Talent Project.

## Ethics declarations

### Competing interests

The authors declare no competing interests.

**Extended Data Fig. 1: Design of the hybrid adaptation module.**

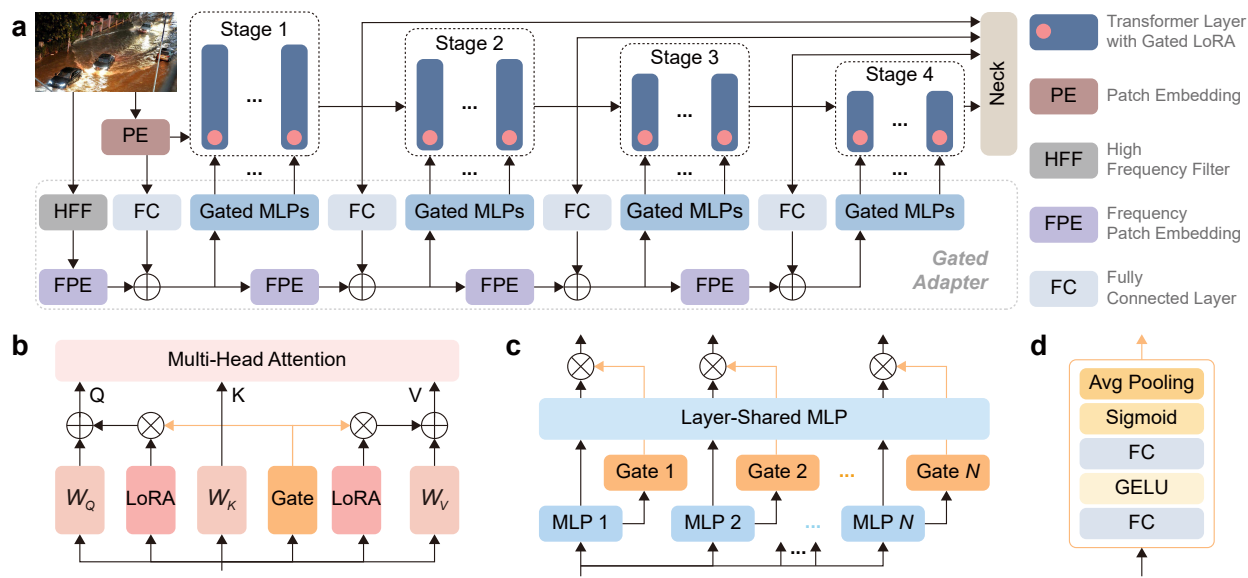

**a** Illustration of the hybrid adaptation module in the SAM2 encoder. Each encoder layer is equipped with a gated Adapter and a gated LoRA module, allowing task-specific adaptation while keeping most pre-trained parameters frozen. **b** Illustration of the Gated LoRA module within the attention module in a Transformer layer. Low-rank matrices are added alongside the original attention weights of query and value, and a gating unit adaptively learns a weight to control the strength of LoRA-based adaptation, ensuring flexible fine-tuning of the attention module. **c** Illustration of the Gated MLPs module in the Gated Adapter. Task-specific information is injected into each Transformer layer through a combination of unshared and layer-shared MLPs, and a gating unit is used to determine the contribution of the Adapter outputs. **d** Detailed design of the gating unit. Given input features, the unit outputs a weight between 0 and 1 that modulates the effect of each adaptation module, thereby harmonizing the Adapter and LoRA for optimal fine-tuning effect.

**Extended Data Fig. 2: Illustration of the proposed S2Match semi-supervised learning method.**

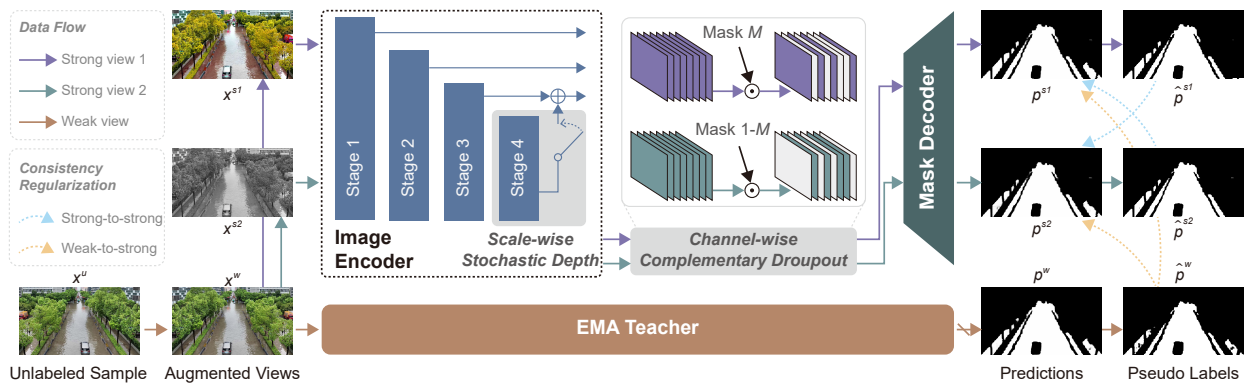

The proposed method belongs to the consistency regularization based semi-supervised learning method. It combines image-level and feature-level augmentations with joint consistency regularization to fully exploit unlabeled data. At the image level, each unlabeled sample is transformed into one weakly augmented view and two strongly augmented views. At the feature level, two complementary strategies are applied: scale-wise stochastic depth randomly drops low-resolution encoder paths to enlarge the structural perturbation space, while channel-wise complementary dropout applies different dropout masks to paired strong views, ensuring diverse and non-overlapping feature perturbations. Consistency regularization is enforced in two forms: 1) weak-to-strong consistency between weak and strong views and 2) strong-to-strong consistency between the two strongly augmented views, encouraging the model to learn perturbation-invariant representations. The teacher model is updated by an exponential moving average (EMA), providing stable pseudo-labels for unlabeled data. This design enables broader exploration of feature perturbations, maximizes the utility of unlabeled data and enhances model robustness and generalization under challenging urban waterlogging scenarios.

**Extended Data Fig. 3: Illustration of the proposed S3CoT prompting strategy.**

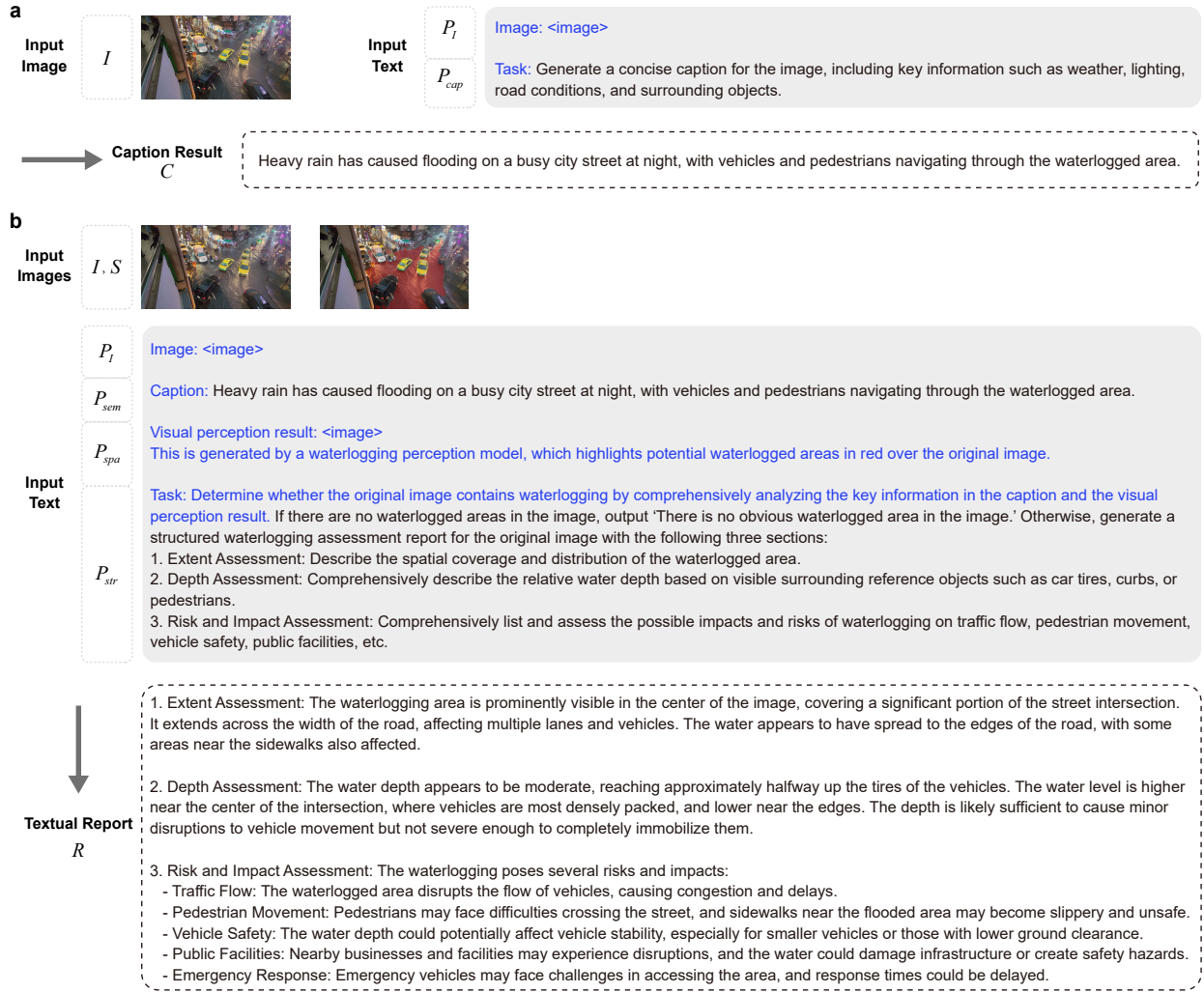

**a** Step 1 of the S3CoT strategy: capturing coarse-grained scene semantics through a captioning task. The input image and task instruction are processed to produce a scene-level caption, which provides semantic context such as weather, lighting, road conditions and surrounding objects. **b** Step 2 of the S3CoT strategy: generating a structured textual report. The caption from Step 1, the visual perception output and the designed reporting instruction are further processed and reformulated into three complementary prompts, i.e., a semantic prompt providing coarse scene semantics, a spatial prompt providing fine-grained localization and a structural prompt providing formatting guidance and reasoning order for the report. In both **a** and **b**, the text highlighted in blue indicates the reformatted or adjusted components that differ from the original input, illustrating how to transform the raw inputs into suitable prompts for the vision–language foundation model. This two-step prompting pipeline ensures the model integrates semantic, spatial and structural information, enabling comprehensive and reliable report generation.

**Extended Data Fig. 4: Illustration on the textual reference report construction process.**

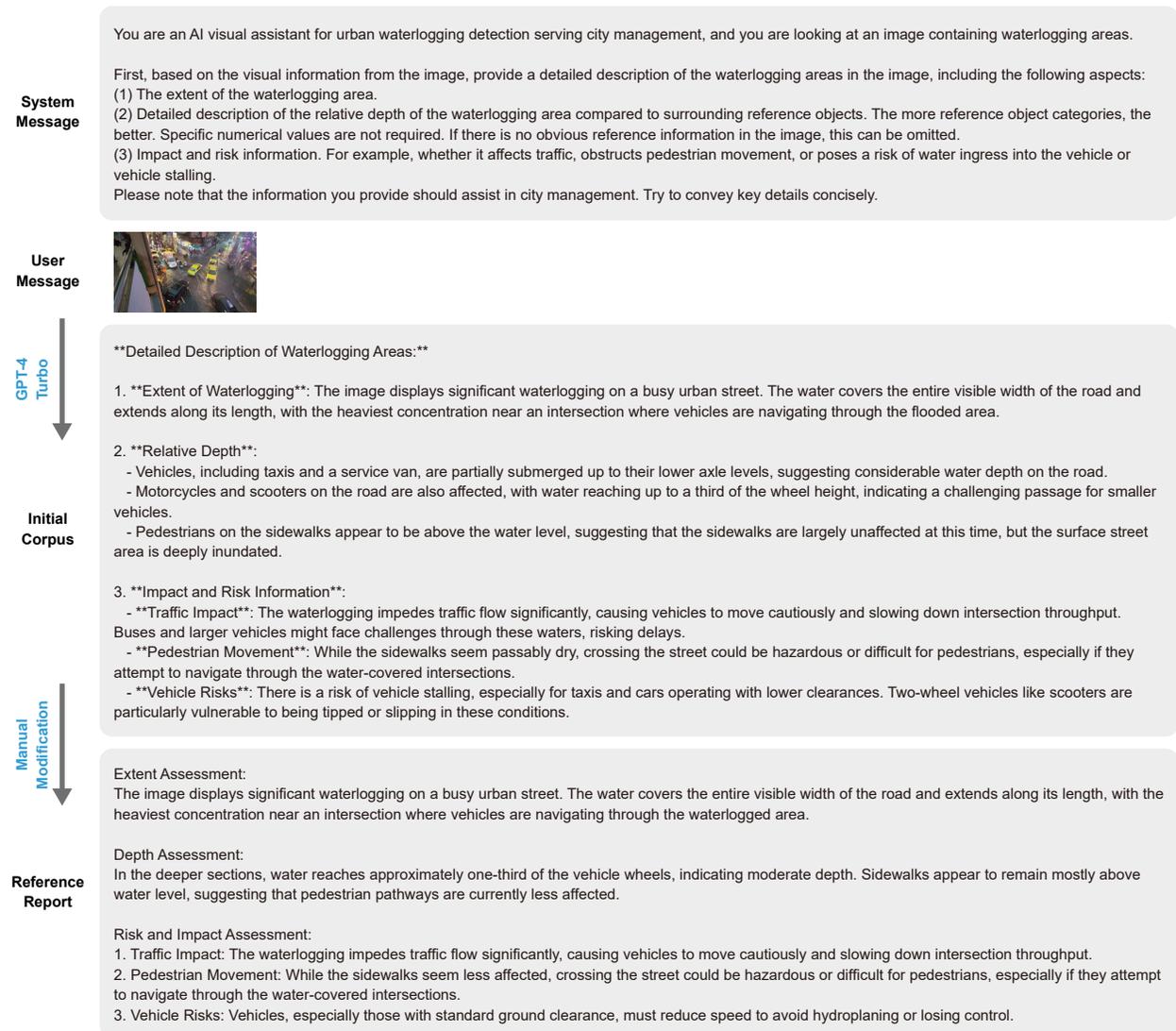

GPT-4 Turbo is prompted with a System Message containing carefully designed textual instructions and a User Message including the input image. The response serves as the initial corpus, which is then manually modified to remove hallucinations and correct factual errors, in order to acquire high-quality reference reports involving extent, depth, risk and impact assessment.

**Extended Data Fig. 5: Textual report evaluation process using GPT-4 Turbo.**

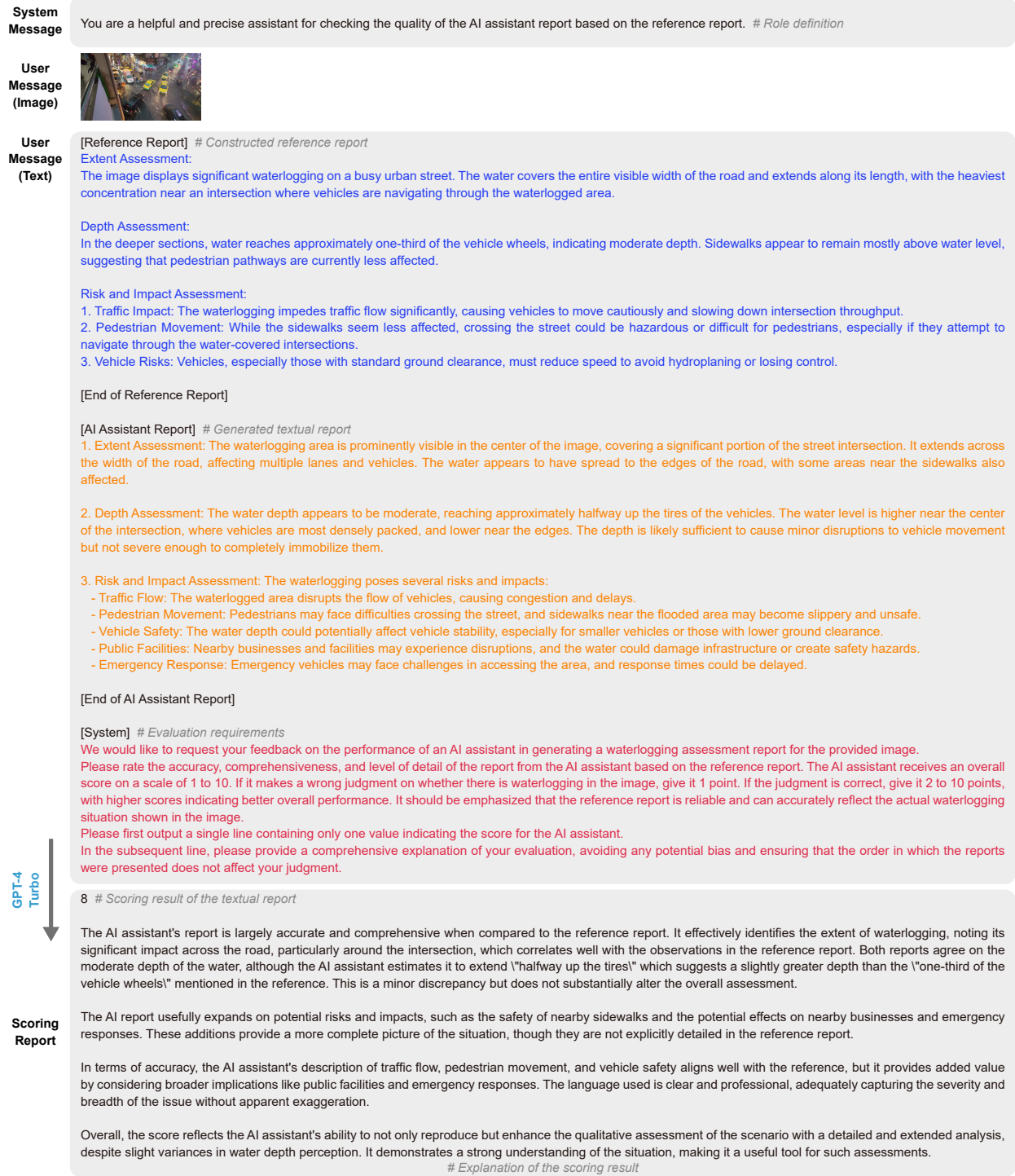

The complete textual report evaluation process. For each test case, GPT-4 Turbo receives a System Message defining its role as an evaluator, a User Message (image) containing the test image, and a User Message (text) containing the reference report (highlighted in blue), the generated report to be evaluated (highlighted in orange) and the evaluation requirements (highlighted in red). The model finally outputs a scoring report, containing an evaluation score and a comprehensive explanation covering accuracy, comprehensiveness and details, etc.

# Supplementary Information

## Supplementary Table 1 Visual perception performance comparisons between different methods

| Method | | U2Net | LSM-Adapter-U | SAM2-Adapter | SAM2-LoRA | Semi-SAM2-Adapter | Semi-SAM2-LoRA | UWAssess |
|---|---|---|---|---|---|---|---|---|
| **UWBench-All** | Precision | 0.7485 | 0.7612 | 0.8136 | 0.7921 | 0.8606 | 0.8715 | 0.9178 |
| | Recall | 0.8232 | 0.8591 | 0.8674 | 0.8825 | 0.9289 | 0.9302 | 0.9473 |
| | Specificity | 0.9435 | 0.9450 | 0.9594 | 0.9527 | 0.9693 | 0.9720 | 0.9827 |
| | Dice | 0.7841 | 0.8072 | 0.8397 | 0.8349 | 0.8934 | 0.8999 | 0.9323 |
| | IoU | 0.6448 | 0.6767 | 0.7236 | 0.7166 | 0.8074 | 0.8180 | 0.8732 |
| | G-Mean | 0.8813 | 0.9010 | 0.9123 | 0.9169 | 0.9489 | 0.9509 | 0.9648 |
| **UWBench-Hard** | Precision | 0.5685 | 0.5806 | 0.5817 | 0.5494 | 0.7255 | 0.6886 | 0.8357 |
| | Recall | 0.6762 | 0.6901 | 0.6912 | 0.7256 | 0.8737 | 0.8580 | 0.8864 |
| | Specificity | 0.9454 | 0.9469 | 0.9471 | 0.9366 | 0.9648 | 0.9587 | 0.9814 |
| | Dice | 0.6177 | 0.6306 | 0.6318 | 0.6253 | 0.7927 | 0.7641 | 0.8603 |
| | IoU | 0.4468 | 0.4605 | 0.4617 | 0.4548 | 0.6566 | 0.6182 | 0.7549 |
| | G-Mean | 0.7995 | 0.8084 | 0.8091 | 0.8244 | 0.9181 | 0.9070 | 0.9327 |
| **Roadway Flooding** | Precision | 0.8957 | 0.9031 | 0.9353 | 0.9477 | 0.9512 | 0.9491 | 0.9582 |
| | Recall | 0.9056 | 0.9153 | 0.9351 | 0.9198 | 0.9192 | 0.9273 | 0.9266 |
| | Specificity | 0.9259 | 0.9310 | 0.9545 | 0.9643 | 0.9669 | 0.9650 | 0.9716 |
| | Dice | 0.9006 | 0.9091 | 0.9352 | 0.9335 | 0.9349 | 0.9381 | 0.9422 |
| | IoU | 0.8192 | 0.8334 | 0.8783 | 0.8753 | 0.8778 | 0.8834 | 0.8907 |
| | G-Mean | 0.9157 | 0.9231 | 0.9448 | 0.9418 | 0.9427 | 0.9460 | 0.9489 |

**Supplementary Table 2 Performance and parameter efficiency of UWAssess under different encoder fine-tuning ratios.**

| Ratio | UWBench-All | | | | | | UWBench-Hard | | | | | | Fine-tuned Paras |
|---|---|---|---|---|---|---|---|---|---|---|---|---|---|
| | Precision | Recall | Specificity | Dice | IoU | G-Mean | Precision | Recall | Specificity | Dice | IoU | G-Mean | |
| 0% | 0.7961 | 0.9131 | 0.9523 | 0.8506 | 0.7400 | 0.9325 | 0.6276 | 0.8152 | 0.9485 | 0.7092 | 0.5494 | 0.8793 | 3.8M |
| 25% | 0.8665 | 0.9332 | 0.9707 | 0.8986 | 0.8159 | 0.9517 | 0.6850 | 0.8652 | 0.9577 | 0.7646 | 0.6190 | 0.9102 | 4.1M |
| 50% | 0.8781 | 0.9484 | 0.9731 | 0.9119 | 0.8381 | 0.9607 | 0.7242 | 0.8834 | 0.9642 | 0.7959 | 0.6610 | 0.9229 | 4.3M |
| 100% | 0.9178 | 0.9473 | 0.9827 | 0.9323 | 0.8732 | 0.9648 | 0.8357 | 0.8864 | 0.9814 | 0.8603 | 0.7549 | 0.9327 | 4.5M |

**Supplementary Table 3 Ablation studies on hybrid adaptation module**

| Tuning Settings | | | | UWBench-All | | | | | | UWBench-Hard | | | | | | Fine-tuned Paras |
|---|---|---|---|---|---|---|---|---|---|---|---|---|---|---|---|---|
| Decoder | Adapter | LoRA | Gate | Precision | Recall | Specificity | Dice | IoU | G-Mean | Precision | Recall | Specificity | Dice | IoU | G-Mean | |
| √ | | | | 0.7961 | 0.9131 | 0.9523 | 0.8506 | 0.7400 | 0.9325 | 0.6276 | 0.8152 | 0.9485 | 0.7092 | 0.5494 | 0.8793 | 3.80M |
| √ | √ | | | 0.8840 | 0.9082 | 0.9757 | 0.8959 | 0.8115 | 0.9413 | 0.7305 | 0.7856 | 0.9691 | 0.7571 | 0.6091 | 0.8726 | 3.86M |
| √ | | √ | | 0.8726 | 0.9459 | 0.9718 | 0.9078 | 0.8312 | 0.9588 | 0.7204 | 0.8777 | 0.9637 | 0.7913 | 0.6547 | 0.9197 | 3.90M |
| √ | √ | √ | | 0.8720 | 0.9430 | 0.9717 | 0.9061 | 0.8283 | 0.9572 | 0.6974 | 0.8880 | 0.9590 | 0.7813 | 0.6410 | 0.9228 | 3.96M |
| √ | √ | √ | √ | 0.9178 | 0.9473 | 0.9827 | 0.9323 | 0.8732 | 0.9648 | 0.8357 | 0.8864 | 0.9814 | 0.8603 | 0.7549 | 0.9327 | 4.54M |

√ indicates fine-tuning training for the corresponding network component.

**Supplementary Table 4** Ablation studies on the threshold for selecting reliable pseudo labeled data in strong-to-strong consistency loss computation

| Threshold | UWBench-All | | | | | | UWBench-Hard | | | | | |
|---|---|---|---|---|---|---|---|---|---|---|---|---|
| | Precision | Recall | Specificity | Dice | IoU | G-Mean | Precision | Recall | Specificity | Dice | IoU | G-Mean |
| 0.5 | 0.8595 | 0.9269 | 0.9691 | 0.8919 | 0.8049 | 0.9477 | 0.6322 | 0.8349 | 0.9483 | 0.7196 | 0.5620 | 0.8898 |
| 0.65 | 0.8823 | 0.9142 | 0.9751 | 0.8980 | 0.8149 | 0.9442 | 0.7104 | 0.8430 | 0.9634 | 0.7710 | 0.6274 | 0.9012 |
| 0.8 | 0.9178 | 0.9473 | 0.9827 | 0.9323 | 0.8732 | 0.9648 | 0.8357 | 0.8864 | 0.9814 | 0.8603 | 0.7549 | 0.9327 |
| 0.95 | 0.9060 | 0.9503 | 0.9799 | 0.9276 | 0.8651 | 0.9650 | 0.7997 | 0.8829 | 0.9765 | 0.8393 | 0.7230 | 0.9285 |

**Supplementary Table 5 Ablation studies on S2Match semi-supervised learning method**

| Tuning Settings | | UWBench-All | | | | | | UWBench-Hard | | | | | |
| --- | --- | --- | --- | --- | --- | --- | --- | --- | --- | --- | --- | --- |
| SC | SD | Precision | Recall | Specificity | Dice | IoU | G-Mean | Precision | Recall | Specificity | Dice | IoU | G-Mean |
|  |  | 0.8475 | 0.9296 | 0.9659 | 0.8866 | 0.7964 | 0.9475 | 0.7048 | 0.8381 | 0.9626 | 0.7657 | 0.6203 | 0.8982 |
| √ |  | 0.8851 | 0.9325 | 0.9753 | 0.9082 | 0.8318 | 0.9537 | 0.7393 | 0.8388 | 0.9685 | 0.7859 | 0.6473 | 0.9013 |
|  | √ | 0.8526 | 0.9395 | 0.9668 | 0.8940 | 0.8082 | 0.9531 | 0.6580 | 0.8952 | 0.9505 | 0.7585 | 0.6109 | 0.9224 |
| √ | √ | 0.9178 | 0.9473 | 0.9827 | 0.9323 | 0.8732 | 0.9648 | 0.8357 | 0.8864 | 0.9814 | 0.8603 | 0.7549 | 0.9327 |

√ indicates that the corresponding component in S2Match is adopted. SC stands for strong-to-strong consistency design and SD stands for stochastic depth design.

**Supplementary Fig. 1: Visual perception performance comparisons.**

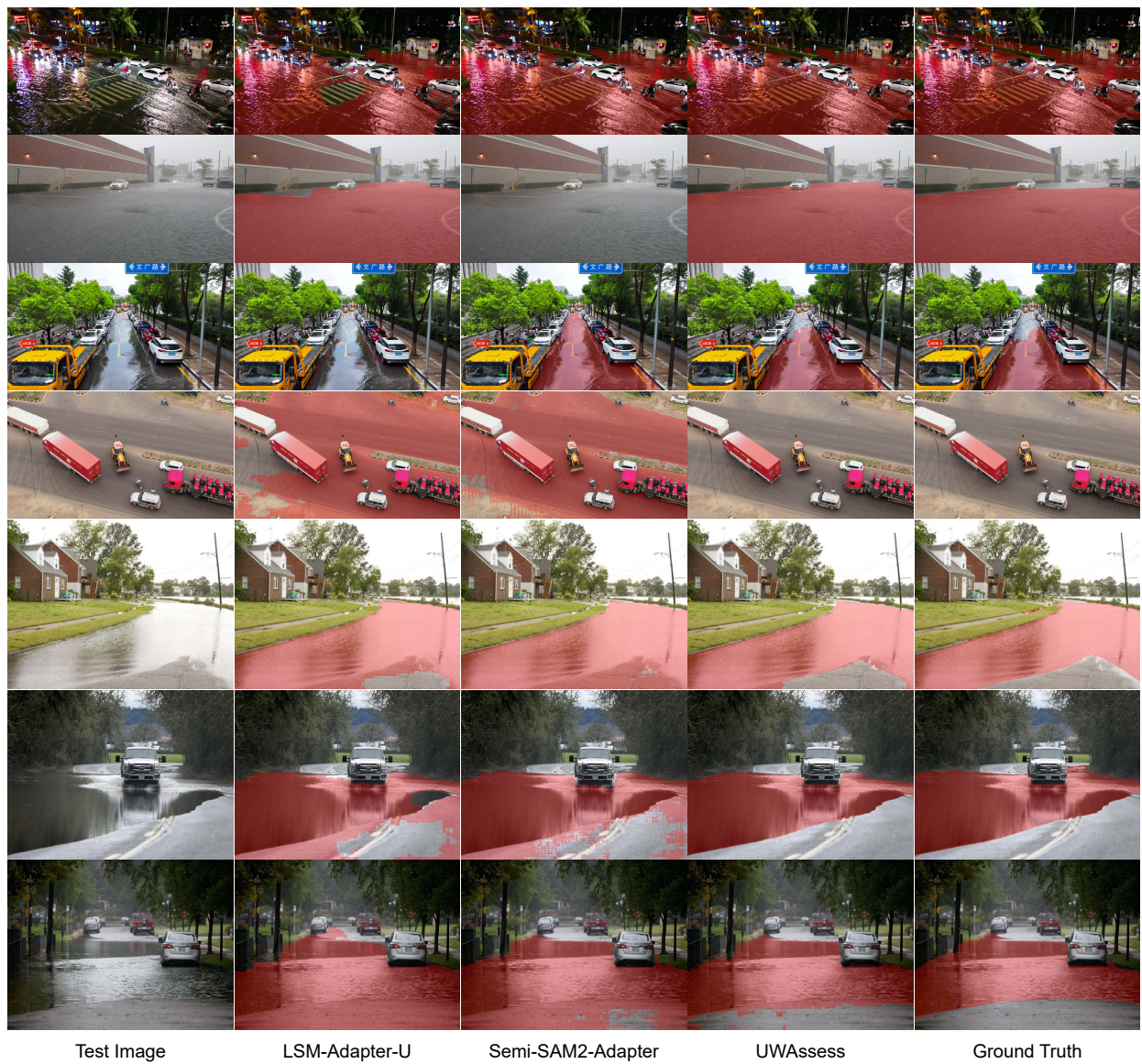

UWAssess obtains reliable visual perception performance across diverse waterlogging appearances, weather conditions, lighting environments, road surface states and camera viewpoints.

**Supplementary Fig. 2: Effectiveness of the structural prompt in textual report generation.**

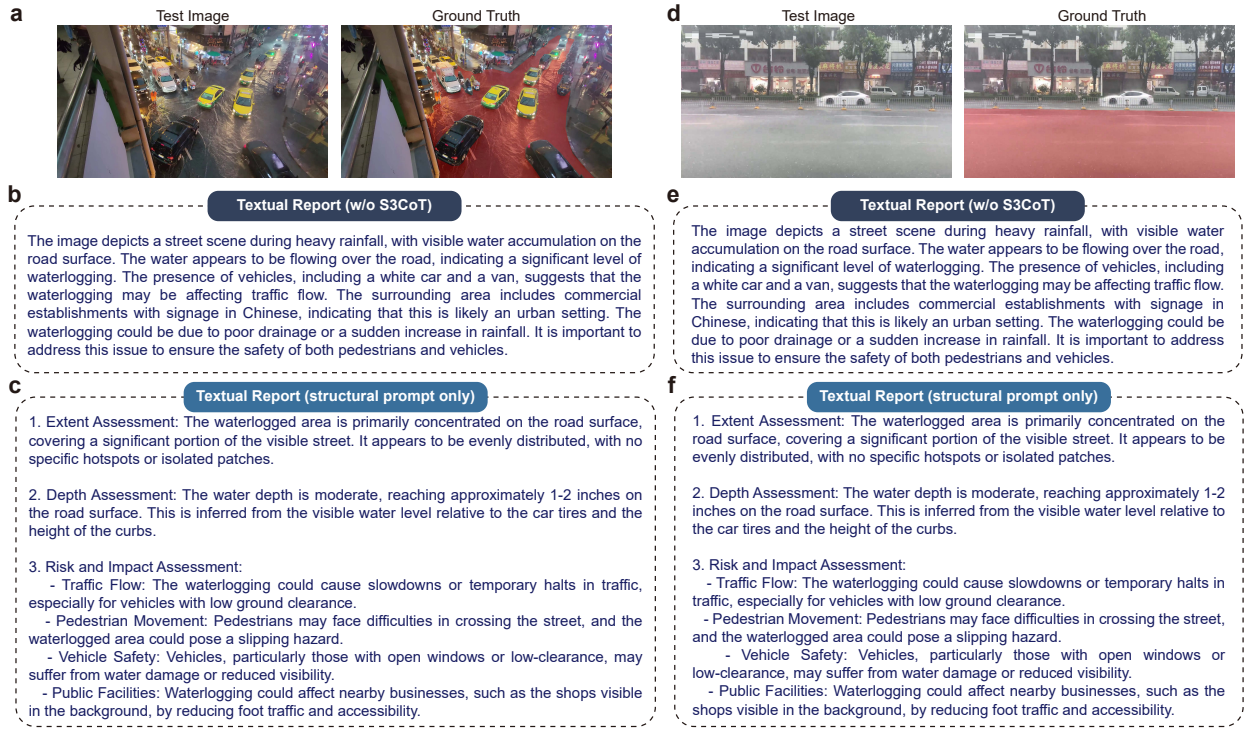

**a, d** Test image and corresponding ground truth mask. **b, e** Textual report generated by UWAssess without using the S3CoT prompting strategy. The generated textual report lacks structural coherence and omits critical information. **c, f** Textual report generated by UWAssess using the structural prompt only. The generated textual assessment report is accurate, comprehensive and well-organized, covering important information such as the extent, depth, and potential risks and impacts of waterlogging observed in the input scene. The results show that the structural prompt provides formatting guidance and reasoning order for the report generation, thereby enhancing the comprehensiveness and readability of the report.